\documentclass[letterpaper,journal]{IEEEtran}
\usepackage{amsmath,amssymb,amsfonts}
\usepackage{bm}
\usepackage{tabularx}
\usepackage{graphicx}
\usepackage{textcomp}
\usepackage{xcolor}
\usepackage{color,soul}
\usepackage{subcaption}
\usepackage[ font=footnotesize]{caption}
\usepackage{url}
\usepackage{booktabs}
\usepackage{lipsum}
\usepackage{gensymb}
\usepackage{algorithm}
\usepackage{algpseudocode}
\usepackage{multicol}
\usepackage{amssymb}
\usepackage{dirtytalk}

\usepackage{algpseudocodex}
\usepackage[normalem]{ulem}
\definecolor{ethblue}{RGB}{54,122,199}

\algnewcommand{\algorithmicforeach}{\textbf{for each}}
\algdef{S}[FOR]{ForEach}[1]{\algorithmicforeach\ #1\ \algorithmicdo}

\algnewcommand{\Initialize}[1]{%
  \State \textbf{Initialize:}
  \Statex \hspace*{\algorithmicindent}\parbox[t]{0.935\linewidth}{\raggedright #1}
}

\newtheorem{rmk}{\textbf{\textit{Remark}}}
\usepackage{pifont}
\newcommand{\cmark}{\textcolor{black}{\ding{51}}}
\newcommand{\xmark}{\textcolor{black}{\ding{55}}}
\usepackage{enumerate}
\usepackage{cite}
\usepackage{hyperref}
\hypersetup{linkbordercolor=green}
\usepackage[english]{babel}

\usepackage{nicefrac}
\usepackage{flushend}
\usepackage{textcomp}
\usepackage{dsfont}

\usepackage{pifont}

\begin{document}

\title{\LARGE \bf
Active Interaction-Aware Model Predictive Path Integral via Ego-Conditioned Generative Predictions
}
\author{Khaled A. Mustafa$^{1*}$, Mohamed-Khalil Bouzidi$^{2*}$, Christian Schlauch$^{2,3}$, Ahmad Gazar$^{1}$, Nadja Klein$^{3}$, Joerg Reichardt$^{2}$, Javier Alonso-Mora$^{1}$
\thanks{This research was supported by funding from the Dutch Research Council NWO-NWA, within the “Acting under uncertainty” (ACT) project (Grant No. NWA.1292.19.298) and  by the German Federal Ministry for Economic Affairs and Climate Action within the project ''NXT GEN AI METHODS''. }
\thanks{$^*$These authors have contributed equally.}
\thanks{$^{1}$Cognitive Robotics Department, TU Delft, The Netherlands, {\tt\small \{k.a.mustafa, j.alonsomora\}@tudelft.nl}}
\thanks{$^{2}$Aumovio SE, Germany, $^{3}$Karlsruhe Institute of Technology, Germany. }
}


\markboth{}%
{Shell \MakeLowercase{\textit{et al.}}: A Sample Article Using IEEEtran.cls for IEEE Journals}


\maketitle

\begin{abstract}
Dense traffic is inherently interactive. The ego vehicle and surrounding agents continuously influence each other's reactions, making  ``what-if'' reasoning essential for safe and efficient driving. To enable such an active interaction-aware behavior, we propose a planning framework that integrates an ego-conditioned generative autoregressive prediction model within Model Predictive Path Integral (MPPI) control. The generative prediction model outputs stochastic, multi-modal predictions of surrounding agents conditioned on each of the ego’s considered future actions. A nested sampling scheme enables tractable evaluation of expected cost and collision risk under the induced distribution. 
This formulation  allows the ego to \textit{actively} probe how different candidate actions shape the interaction outcomes and to identify actions that reduce ambiguity in uncertain interactions. Closed-loop simulations demonstrate improved safety and efficiency compared to conventional predict-then-plan and passive interaction-aware approaches. 
\end{abstract}

\section{Introduction}
\IEEEPARstart{A}{utonomous} vehicles are required to navigate in complex, interactive scenarios involving heterogeneous agents, such as cooperative and non-cooperative human drivers alongside other autonomous vehicles. A fundamental challenge, in these environment, arises from the inherent coupling between prediction and planning. The future behavior of surrounding agents depends on the ego vehicle's actions, while safe and efficient planning relies on accurately anticipating how those agents will behave. Traditional predict-then-plan approaches, e.g., \cite{sh-mpc, 25-dra, 24-cdc}, decouple these processes, first predicting agent trajectories independently of the ego vehicle's planned motion, and then planning based on these fixed predictions. However, this paradigm inevitably suffers from the well-known \textit{frozen robot problem} \cite{frp}. By failing to account for the ego vehicle's influence, the system exhibits excessively cautious or deadlocked behavior in interactive scenarios like dense merging, unprotected left turns, or narrow passages.

In contrast, human drivers naturally resolve such scenarios through active interaction \cite{Sadigh2016, Sadighrss, Schwarting}. They implicitly understand that their actions provoke reactions and use this to probe the hidden intent (e.g., cooperativeness) of other drivers. For example, inching forward during a merge solicits a reaction, revealing whether nearby vehicles will yield or maintain speed. By taking an action specifically to observe a reaction, the driver narrows their belief about what the other agents will do. This process of \textit{active uncertainty reduction} enables humans to efficiently navigate ambiguous interactions without being overly aggressive or excessively conservative.
The recent success of generative models for motion generation trained on large-scale  real-world data presents a promising opportunity to endow autonomous vehicles with similar interactive capabilities. In particular, autoregressive transformer-based architectures such as SMART \cite{smart}, MotionLM \cite{motionlm}, or Trajeglish \cite{Trajeglish} have emerged as powerful tools. The key strength  is their autoregressive factorization. By predicting agent futures step-by-step through next-token prediction, they naturally capture temporal relations in the joint distribution over agent trajectories. 
This sequential structure is particularly well suited for modeling interactions. It allows the model to condition each agent’s next state on the set of interacting agents, including, critically, the planned trajectory of the ego vehicle.
Moreover, their stochastic sampling mechanism enables diverse yet realistic rollouts, making them ideal for representing the multi-modal uncertainty inherent in interactive traffic scenarios.

In this work, we leverage SMART for ego-conditioned prediction, to tackle the underlying problem of the frozen robot \cite{frp}. However, ego-conditioned predictions introduce a circular dependency: the ego vehicle requires a prediction to plan, but the prediction depends on the specific plan being executed over time.
To address this, we integrate ego-conditioned prediction within the MPPI framework \cite{mppi_williams}. MPPI provides a principled way to resolve the circular dependency between prediction and planning. This stems from the fact that MPPI naturally samples a diverse set of candidate ego trajectories through control perturbations. For each hypothesized plan, SMART acts as an interactive simulator, updating its rollouts and beliefs about surrounding agents in response to the candidate ego states. By evaluating the safety and efficiency costs of these  rollouts, the planner assesses the resulting distribution over interaction outcomes. This enables exploration of possible ego maneuvers and their corresponding effects on surrounding agents, allowing the planner to balance task objectives against interaction uncertainty.

\section{Related Work}
\textcolor{black}{While trajectory prediction has advanced significantly with generative models, effectively integrating these forecasts into closed-loop planners for interactive settings remains a fundamental challenge. Existing literature addresses this interaction, specifically how to model the ego vehicle's influence on surrounding agents, through four broad paradigms:}
\subsubsection{Multi-modal uncertainty-aware planner} A common way to exploit predictions is through a \textit{predict-then-plan} paradigm, where the ego agent first generates probabilistic multi-modal forecasts of other agents’ motions and subsequently plans a trajectory under these predictions. This includes contingency-based approaches using multi-modal trajectory predictions \cite{ racp, b-mpc}, risk-aware planners that optimize expected cost under predicted distributions \cite{23-mustafa, 24-cdc}, and chance-constraint formulations that enforce probabilistic safety guarantees \cite{s-mpc, cc}. While these methods account for uncertainty in others' future motion, they ignore that the ego's actions influence surrounding agents. As a result, they often exhibit overly conservative behavior in interactive scenarios and suffer from the frozen robot problem \cite{frp}. 
\subsubsection{Game-theoretic interaction-aware planner} Recognizing this limitation, a second line of work models the interaction explicitly by formulating motion planning as a joint optimization problem over multiple agents. Game-theoretic approaches provide a principled framework for this setting by treating agents as rational decision-makers and casting interaction as a dynamic game \cite{lecleach_algames_2022, 24-peters, BOUZIDI2025104908}. However, these methods suffer from two major drawbacks. First, computational complexity grows rapidly with the number of agents, rendering the equilibrium computation intractable in dense traffic scenes. Second, the assumption of rational, utility-maximizing behavior is often violated in practice, particularly when interacting with human drivers whose actions may be inconsistent, boundedly rational, or influenced by latent factors not captured by the model.
\subsubsection{Ego-conditioned interaction-aware planner}
To balance expressiveness and tractability, ego-conditioned interaction-aware planning has emerged as an intermediate paradigm. In this setting, the predictions of other agents’ behavior are conditioned on candidate ego trajectories \cite{dtpp, Chen2023PPADII, iann-mppi}. By conditioning predictions on ego actions, these approaches acknowledge that other agents react to the ego vehicle's behavior and behave more assertively than predict-then-plan methods. However, this conditioning imposes a leader-follower structure, i.e., the ego acts, others react. This may bias the planner toward overly aggressive behavior, as it implicitly expects other agents to yield or adapt to the ego’s actions to avoid collision.
\subsubsection{Active dual-control-based planner}
Unlike the previously mentioned methods, dual control and information-theoretic approaches offer a more principled alternative by \emph{actively} reducing uncertainty \cite{Sadigh2016, 24-fisac, dual_mppi}. These methods explicitly reason about the epistemic uncertainty in other agents' predicted behaviors by formulating planning in belief space. This enables active \emph{uncertainty reduction}, meaning the ego vehicle actively probes other agents to resolve uncertainty about their intent, for instance w.r.t. the cooperativeness of other drivers. \emph{Implicit} dual control formulations \cite{24-fisac, dual_mppi} are particularly attractive, as they avoid the need for an explicit information-gain term. Instead, the exploration-exploitation trade-off emerges naturally from uncertainty propagation in the stochastic optimization. While theoretically appealing, such methods  typically rely on simplified parametric models of other agents, which can fail to capture complex interaction dynamics, diversity in real-world human driving behavior. As a result, such methods are often limited to idealized scenarios and exhibit reduced behavioral expressiveness. Table~\ref{tab:comparison} summarizes the comparison between our method and the previously discussed approaches. 

\subsection{Statement of Contributions}

\begin{enumerate}[(i)]
\item We propose a hybrid planning framework that integrates learned autoregressive generative  models for multi-agent prediction into
    MPPI,  closing the loop between prediction and planning. Unlike dual-control methods, our approach leverages the expressive capacity of generative model to capture multi-modal uncertainty from real-world data while retaining the model-based  safety-relevant 
    structure.
    \item 
    We introduce an interaction- and uncertainty-aware planning formulation using a tractable sampling scheme to evaluate multi-modal predictions conditioned on sampled ego trajectories  yielding risk-aware collision probability estimates. This prevents overly aggressive leader-follower bias of standard ego-conditioned planners.
    \item    We show that combining the generative model with the MPPI weighting mechanism enables active uncertainty reduction without requiring explicit belief-space representations or dedicated exploration terms, and that it yields more assertive, efficient driving.

\end{enumerate} 

\begin{figure*}[h!]
    \medskip
  \centering  \includegraphics[width=\textwidth]{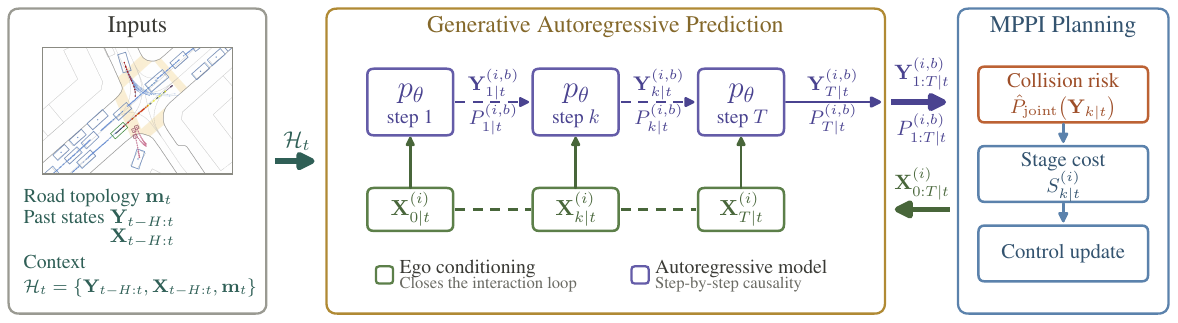}
  \caption{Overview of the proposed framework integrating MPPI with an generative autoregressive prediction model which rolls out surrounding-agent hypotheses step by step, conditioned on the MPPI ego trajectory samples}
  \label{fig:method_overview}
\end{figure*}

\begin{table}[h]
\caption{Planner comparison by uncertainty-, interaction-awareness, learned predictions, and active probing. (\cmark) indicates capability is present in some methods of the paradigm, but not jointly with other (\cmark) capabilities.}
\label{tab:comparison}
\centering
\resizebox{\columnwidth}{!}{%
\begin{tabular}{lcccc}
\toprule
Method 
& Uncertainty
& Interaction
& Learning-aided
& Active \\
\midrule
Multi-modal~\cite{racp,24-cdc,s-mpc} 
& \cmark & (\cmark) & (\cmark) & \xmark \\

Game-theoretic~\cite{24-peters,lecleach_algames_2022, BOUZIDI2025104908} 
& (\cmark) & \cmark &  (\cmark) & \xmark  \\

Ego-conditioned~\cite{dtpp, iann-mppi,Chen2023PPADII} 
& \xmark & \cmark & \cmark & \xmark \\

Dual control~\cite{Sadigh2016, 24-fisac, dual_mppi} 
& \cmark & \cmark & \xmark & \cmark  \\

Ours
& \cmark & \cmark & \cmark & \cmark  \\
\bottomrule
\end{tabular}
}
\end{table}

\section{Preliminaries}
We consider motion planning for an autonomous vehicle operating in a dynamic traffic environment with multiple interacting traffic participants, collected in set $\mathcal{A}$. Time is discretized with index $t \in \mathbb{N}$. At time $t$, the ego vehicle state is denoted by $\mathbf{x}_t \in \mathcal{X}$, the control input by $\mathbf{u}_t \in \mathcal{U}$, and the \textit{joint state} of surrounding agents by $\mathbf{Y}_t \in \mathcal{Y}$, where $\mathbf{Y}_t = [\mathbf{y}^{(0)}_t, \cdots, \mathbf{y}_t^{(|\mathcal{A}|)})]^\top $ and  $\mathbf{y^{(a)}}$, $a \in \mathcal{A}$  denotes a single agent's state. The ego vehicle evolves according to known deterministic dynamics
\begin{equation}
\label{model}
\mathbf{x}_{t+1} = \mathcal{F}(\mathbf{x}_t, \mathbf{u}_t),
\qquad \mathbf{x}_t \in \mathcal{X}, \ \mathbf{u}_t \in \mathcal{U},
\end{equation}
whereas the future evolution of the surrounding agents is unknown and treated as stochastic. Planning is performed over a finite horizon $T$. Within a planning iteration at real-world time $t$, future quantities are indexed by a rollout index $k = 1, \dots, T$. The ego state, control input, and surrounding agents' state at rollout step $k$ are denoted by $\mathbf{x}_{k\mid t}$, $\mathbf{u}_{k\mid t}$, and $Y_{k\mid t}$, respectively. The rollouts are initialized using the measured current states,
\begin{equation}
\mathbf{x}_{0\mid t} = \mathbf{x}_t^{\mathrm{meas}}, \qquad
\mathbf{Y}_{0\mid t} = \mathbf{Y}_t^{\mathrm{meas}}.
\end{equation}
For $k > 0$, ego states evolve according to the deterministic dynamics under sampled control inputs, while other agents' states are generated stochastically using a learned prediction model that captures interaction and behavioral uncertainty.

\subsection{Model Predictive Path Integral Control}

Model Predictive Path Integral Control (MPPI) \cite{williams_information-theoretic_2018}, serving as the backbone of this paper, solves finite-horizon stochastic optimal control problems using a zeroth-order, sampling-based Model Predictive Control (MPC) scheme. At time step $t$, MPPI samples $M$ random control sequences around a nominal control sequence $\mathbf{U}_{t} = [\mathbf{u}_0, \dots, \mathbf{u}_{T-1}]^\top$ over a horizon $T$. This perturbed sequence for the $i$-th rollout is given by:
\begin{equation}
\mathbf{U}_{t}^{(i)} = \mathbf{U}_{t} + \boldsymbol{\epsilon}^{(i)}, \quad
\boldsymbol{\epsilon}^{(i)}_k \sim \mathcal{N}(\mathbf{0}, \boldsymbol{\Sigma}_{\epsilon}),
\quad i = 1, \dots, M,
\end{equation}
where $\boldsymbol{\epsilon}^{(i)}$ is a randomly sampled control perturbation. These sequences are rolled out through the ego dynamics $\mathcal{F}(\cdot)$ to obtain state trajectories $\mathbf{X}_{t}^{(i)} = \{\mathbf{x}^{(i)}_{k \mid t}\}_{k=0}^{T}$, from which a total trajectory cost $S^{(i)}$ is computed.  The optimal control sequence is then approximated using an exponential importance-sampling update law:
\begin{equation}
\label{eq:mppi_update}
    \mathbf{U}_{t}^{*} = \sum_{i=1}^{M} \frac{1}{\eta}\exp\left(-\frac{1}{\beta}\left(S^{(i)} - \rho\right)\right) \mathbf{U}_{t}^{(i)},
\end{equation}
where $\beta$ is the inverse temperature parameter that determines the selective pressure over samples, $\rho = \min_i S^{(i)}$ is subtracted to maintain numerical stability, and $\eta$ normalizes the weights. Following the receding horizon principle, only the first optimal control input $\mathbf{u}_0^*$ is applied to the system. The entire procedure is then repeated at the next time step, warm-started by a time-shifted version of $\mathbf{U}^*_{t}$.

\section{Methodology}
Our planning framework integrates MPPI with a generative autoregressive prediction model (GARPM).
In the nominal MPPI formulation, the cost depends on the future states
$\mathbf{Y}_{k\mid t}$ of surrounding agents, which are unknown at planning time and can also depend on the ego vehicle’s future motion.  To account for this uncertainty and interaction dependence, we model surrounding agents’ future behavior using the GARPM and condition it on the ego trajectory and the current traffic context. Rather than explicitly parameterizing agent dynamics through latent interaction variables, the model directly generates joint predictions of surrounding agents’ future trajectories.
For each candidate ego trajectory induced by MPPI, we approximate the resulting interaction outcomes by sampling multiple stochastic rollouts from the GARPM. Trajectory costs are then evaluated with respect to the belief the ego-vehicle maintains over the rollouts sampled from this distribution. The following subsections first introduce the underlying active planning formulation and then describe a tractable sampling-based approximation. 
An overview of the proposed method is illustrated in Fig. \ref{fig:method_overview}.

\subsection{Active Planning Formulation}
\label{sec:active_planning}
Due to latent objectives, interaction effects, and unobserved internal states of surrounding agents, their future behavior cannot be predicted deterministically. The evolution of surrounding agents is therefore modeled as a stochastic process conditioned on the observed traffic context. 

At time step $t$, the ego observes the road topology $\mathbf{m}_t \in \mathcal{M}$, dynamic traffic guidance $\mathbf{g}_t \in \mathcal{G}$, and the current joint state of surrounding agents $\mathbf{Y}_t$. Given a context $
\mathcal{H}_t =
\left\{
\mathbf{Y}_{t-H:t-1|t},
\mathbf{X}_{t-H:t-1|t},
\mathbf{m}_t,
\mathbf{g}_t
\right\},
$
which includes observations over the past $H$ time steps for all agents, we model the joint distribution over future agent states over a horizon of length $T$ as
\begin{equation}
p(\mathbf{Y}_{1:T \mid t} \mid \mathbf{Y}_{0 \mid t}, \mathcal{H}_t)
=
p(\mathbf{Y}_{1 \mid t}, \ldots, \mathbf{Y}_{T \mid t}
\mid \mathbf{Y}_{0 \mid t}, \mathcal{H}_t).
\end{equation}

\noindent We employ a generative autoregressive formulation, in which the modeled joint distribution factorizes as
\begin{equation}
\label{eq:ar_factorization}
p_{\theta}\!\left(\mathbf{Y}_{1:T \mid t} \mid \mathbf{Y}_{0 \mid t}, \mathcal{H}_t\right)
=
\prod_{k=1}^{T}
p_{\theta}\!\left(
\mathbf{Y}_{k \mid t} \mid \mathbf{Y}_{0:k-1 \mid t}, \mathcal{H}_t
\right),
\end{equation}
where $\theta$ denotes the learned model parameters.\\
 The future traffic participants behavior depends not only on the history states but also on their uncertain anticipation of the ego behavior. 
Hence, we model the future evolution of surrounding agents over the planning horizon by extending \eqref{eq:ar_factorization} to an ego-conditioned predictive distribution:
\begin{equation}
\begin{aligned}
p_{\theta}\!\Big(
\mathbf{Y}_{1:T \mid t} 
\mid
&\mathbf{X}_{0:T \mid t}, \mathbf{Y}_{0 \mid t}, \mathcal{H}_t
\Big)
=\\
&\prod_{k=1}^{T}
p_{\theta}\!\Big(
\mathbf{Y}_{k \mid t}
\mid
\mathbf{X}_{0:k-1 \mid t},
\mathbf{Y}_{0:k-1 \mid t},
\mathcal{H}_t
\Big),
\end{aligned}
\label{eq:ego_cond_pred}
\end{equation}
where we explicitly provide the ego trajectory $\mathbf{X}_{0:T \mid t} = \{\mathbf{x}_{k \mid t}\}_{k=0}^{T}$. The planning objective can be formulated as the expected cost under both the rollout distribution $\mathbb{Q}$ over ego trajectories and the ego-conditioned predictive distribution of surrounding agents $p_\theta(\mathbf{Y}_{1:T \mid t}\mid\mathbf{X}_{0:T \mid t}, \mathbf{Y}_{0 \mid t}, \mathcal{H}_t)$:
\begin{equation}
\label{eq:active_objective}
\begin{aligned}
J_t
&=
\mathbb{E}_{\mathbf{U}_{t} \sim \mathbb{Q}}
\Big[
\mathbb{E}_{\mathbf{Y}_{1:T \mid t} \sim p_\theta(\mathbf{Y}_{1:T \mid t} \mid \mathbf{X}_{0:T \mid t}, \mathbf{Y}_{0 \mid t}, \mathcal{H}_t)}
\big[
\phi(\mathbf{x}_{T \mid t}, \mathbf{Y}_{T \mid t})
\\
&\quad\quad\quad\quad\quad
+ \sum_{k=0}^{T-1}
\ell(\mathbf{x}_{k \mid t}, \mathbf{u}_{k\mid t}, \mathbf{Y}_{k \mid t})
\big]
\Big],
\end{aligned}
\end{equation}
where $\ell(\cdot)$ and $\phi(\cdot)$ denote the stage and terminal cost functions,
respectively. Note that $\mathbf{X}_{0:T\mid t}$ is not a free variable in the inner
expectation: it is fully determined by the sampled control sequence
$\mathbf{U}_t \sim \mathbb{Q}$ and the ego dynamics, $\mathbf{x}_{k+1\mid t} = f(\mathbf{x}_{k\mid t}, \mathbf{u}_{k\mid t})$. This dependence is what couples the outer and inner expectations, making the predictive distribution itself a function of the ego trajectory being optimized.
\begin{rmk}
    \textit{
The optimal solution to this formulation is considered \emph{active}.  Different ego trajectories induce different predictive distributions, yielding varying levels of uncertainty in the interaction which directly affect the expected cost. Minimizing the cost therefore implicitly balances task-specific objectives against active uncertainty reduction.
}
\end{rmk}

\subsection{Sampling-Based Approximation}
Direct evaluation of the objective in~\eqref{eq:active_objective} is computationally
intractable. The predictive distribution over surrounding agents is high-dimensional,
multi-modal, and requires approximation. Moreover, it depends on
the ego trajectory being optimized, resulting in a circular dependency. To obtain a tractable approximation, we employ a nested sampling strategy. The outer
expectation over ego trajectories is approximated using $M$ MPPI rollouts. To approximate the inner expectation over surrounding agents’ state, we draw, for each ego rollout, $B$ samples from the ego-conditioned GARPM. 

\subsubsection{Ego-Conditioned Autoregressive Sampling}
For each of the $M$ MPPI-sampled ego trajectories $\{\mathbf{X}_{0:T|t}^{(i)}\}_{i=1}^{M}$, we generate $B$ stochastic behavior rollouts from the ego-conditioned predictive distribution~\eqref{eq:ego_cond_pred}. To generate these rollouts, we employ a downsized variant of SMART~\cite{smart}.

Specifically, we must evaluate the single-step transition probability that appears inside the temporal product of \eqref{eq:ego_cond_pred}. Under autoregressive motion models, the next \textit{token} for every agent at each prediction step $k$ is decoded in parallel by a transformer that processes the joint previous prediction step of all agents as shared context.
 The structural assumption is that inter-agent influence is captured entirely through the conditioning on past prediction steps rather than within the current decoding step. The joint one-step predictive distribution, therefore, factorizes over agents, as
\begin{equation}\label{eq:facdis}
\begin{aligned}
p_{\theta}\!\Big(
\mathbf{Y}_{k \mid t}^{(i,b)}
\mid 
&\mathbf{Y}^{(i,b)}_{0:k-1 \mid t},
\mathbf{X}^{(i)}_{0:k-1 \mid t},
\mathcal{H}_t
\Big)
=\\
&\prod_{a \in \mathcal{A}}
p_{\theta}\!\left(
\mathbf{y}^{(i,b,a)}_{k \mid t}
\mid
\mathbf{Y}^{(i,b)}_{0:k-1 \mid t},
\mathbf{X}^{(i)}_{0:k-1 \mid t},
\mathcal{H}_t
\right),
\end{aligned}
\end{equation}
indexed by superscripts $i$, $b$, and $a$ for the 
MPPI sample, behavior rollout, and agent, respectively
The joint distribution remains coupled across agents through the shared conditioning context; this is distinct from assuming marginal independence between agents, as interactions are fully captured in the context and updated step by step as the rollout proceeds. Using the factored formulation in \eqref{eq:facdis}, each rollout $b \in \{1,\ldots,B\}$ is sampled autoregressively. At each prediction timestep $k$, we draw a token for each individual agent as
\begin{equation}
    \mathbf{y}_{k|t}^{(i,b,a)} \sim p_\theta\!\left(\,\mathbf{y}_{k \mid t}^{(a)} \,\big|\, \mathbf{Y}_{0:k-1|t}^{(i,b)},\, \mathbf{X}_{0:k-1|t}^{(i)},\, \mathcal{H}_t\right),
    \label{eq:ar_sampling}
\end{equation}
with the shared initial condition $\mathbf{Y}_{0|t}^{(i,b)} = \mathbf{Y}_t^{\mathrm{meas}}$. To balance computational efficiency with coverage of multi-modal behavior, we employ top-$K$ sampling~\cite{smart}, restricting the support at each autoregressive step to the $K$ most likely tokens. This preserves the dominant behavioral modes while avoiding the computational burden of exhaustive Monte Carlo sampling over the high-dimensional distribution.

To subsequently evaluate the probability of these generated trajectories, the likelihood of each per-agent rollout up to timestep $k$ is computed recursively as
\begin{equation}
    \pi_{k|t}^{(i,b,a)} = \pi_{k-1|t}^{(i,b,a)} \cdot p_\theta\!\left(\mathbf{y}_{k|t}^{(i,b,a)} \,\big|\, \mathbf{Y}_{0:k-1|t}^{(i,b)},\, \mathbf{X}_{0:k-1|t}^{(i)},\, \mathcal{H}_t\right),
    \label{eq:ar_likelihood}
\end{equation}
quantifying the plausibility of each sampled hypothesis under the learned model.

\subsubsection{Gaussian Mixture Surrogate Distribution}
The $B$  behavior hypotheses $\{\mathbf{Y}^{(i,b)}_{1:T \mid t}\}_{b=1}^{B}$ provide a sparse, discrete approximation of  $p_{\theta}(\mathbf{Y}_{0:T \mid t} \mid \mathbf{X}^{(i)}_{0:T \mid t}, \mathbf{Y}_{0 \mid t}, \mathcal{H}_t)$. While SMART can represent arbitrary nonlinear distributions, these finite samples do not necessarily correspond to the distribution's modes. Identifying true modes would require exhaustive Monte Carlo sampling followed by kernel density estimation which is prohibitive for real-time operation. To address this, we convert the $B$ discrete samples into a smooth, continuous density function by constructing a surrogate Mixture of Gaussians (MoG).  This surrogate directly integrates the results of our sampling and likelihood steps: we treat each generated agent position from~\eqref{eq:ar_sampling} as the spatial mean of a Gaussian component, and we use the recursive sequence likelihoods from~\eqref{eq:ar_likelihood} to determine that component's mixing weight. This is justified
when $B$ is chosen large enough that the top-$K$ sampling procedure provides adequate coverage of the
dominant behavioral modes. This  naturally captures the multi-modal structure inherent in traffic  behavior \cite{wayformer}. Specifically, for each agent $a$, MPPI trajectory sample $i$, and prediction timestep $k$, the continuous per-agent surrogate distribution is defined as:

\begin{equation}
    \label{eq:gmm}
    \tilde{p}\!\Big(
  \mathbf{y}^{(a)}_{k\mid t}
  \;\mid\;
  \mathbf{X}^{(i)}_{0:k\mid t},   
  \mathcal{H}_t
\Big)
=
\sum_{b=1}^B w_{k\mid t}^{(i,b,a)}
\mathcal{N}\!\left(
  \mathbf{y}_{k\mid t}^{(a)};\,
  \mathbf{y}_{k\mid t}^{(i,b,a)},
  \boldsymbol{\Sigma}
\right),
\end{equation}
where the mean $\mathbf{y}_{k\mid t}^{(i,b,a)}$ is the exact state realization sampled in~\eqref{eq:ar_sampling} for rollout $b$. Because the underlying generative architecture outputs discrete trajectory coordinates rather than explicit variance bounds, we define $\boldsymbol{\Sigma} = \sigma^2 \mathbf{I}$ as a fixed, isotropic covariance matrix. Here, the variance $\sigma^2$ acts as a regularizing hyperparameter that reflects uncertainty around each rollout $b$. The categorical importance weights $w^{(i,b,a)}_{k \mid t}$ are computed by normalizing the rollout likelihoods accumulated up to step $k$ via~\eqref{eq:ar_likelihood}:
\begin{equation}
 w^{(i,b,a)}_{k \mid t}
=
\frac{\pi^{(i,b,a)}_{k \mid t}}
{\sum_{b'=1}^{B} \pi^{(i,b',a)}_{k \mid t}},
\qquad
\sum_{b=1}^{B} w^{(i,b,a)}_{k \mid t} = 1.
\end{equation}
By defining the weights this way, we assign a larger probability mass to spatial regions surrounding hypotheses that the underlying autoregressive model scored as highly plausible. Per-agent weighting is consistent with the factorization in~\eqref{eq:facdis}, where each agent's marginal is evaluated independently. The surrogate is constructed separately at each timestep $k$, to allow the predicted distribution to shift, in both location and relative mode
weights, as new ego-conditioned information accumulates over
the horizon.

\subsubsection{MPPI Evaluation}
Using the surrogate~\eqref{eq:gmm}, the inner expectation in~\eqref{eq:active_objective} is approximated by evaluating stage costs against the hypotheses $B$ and marginalizing with the mixture weights. Given an ego trajectory, the expected cost results in
\begin{equation}
\begin{aligned}
\mathbb{E}_{\mathbf{Y}_{0:T\mid t} \sim p_\theta}&
\!\left[\sum_{k=0}^{T-1} \ell(\mathbf{x}^{(i)}_{k\mid t},
\mathbf{u}^{(i)}_{k\mid t}, \mathbf{Y}_{k\mid t})\right]
\approx \\
&\sum_{k=0}^{T-1}\sum_{b=1}^{B}
w^{(i,b)}_{k\mid t}\,
\ell\!\left(\mathbf{x}^{(i)}_{k\mid t},
\mathbf{u}^{(i)}_{k\mid t},
\mathbf{Y}^{(i,b)}_{k\mid t}\right),
\end{aligned}
\label{eq:approx_inner_exp}
\end{equation}
where $w^{(i,b)}_{k\mid t} = \frac{1}{|\mathcal{A}|}\sum_{a\in\mathcal{A}} w^{(i,b,a)}_{k\mid t}$ aggregates per-agent weights into a joint hypothesis weight. The surrogate~\eqref{eq:gmm} enables a tractable approximation of the collision probability used in the risk component of the stage cost. We model the ego and agent $a$ collision regions, $\mathcal{V}_{k\mid t}$ and $\mathcal{D}^{(a)}_{k\mid t}$, as unions of circles. The marginal collision probability is obtained by integrating~\eqref{eq:gmm} over $\mathcal{R}^{\circ,{(a)}}_{k\mid t}$:
\begin{equation}
\hat{P}\!\left(\mathbf{y}^{(a)}_{k \mid t}\right)
=
\int_{\mathcal{R}^{\circ,a}_{k \mid t}}
\tilde{p}\!\left(\mathbf{y}^{(a)}_{k \mid t}
\mid \mathbf{X}^{(i)}_{0:k\mid t}, \mathcal{H}_t \right)
\, d\mathbf{y}^{(a)}_{k \mid t},
\end{equation}
where $\mathcal{R}^{\circ,{(a)}}_{k\mid t} = \mathcal{V}_{k\mid t} \oplus \mathcal{D}^{(a)}_{k\mid t}$ is the Minkowski sum that inflates the ego footprint by the agent's shape. Each Gaussian component's contribution reduces to a cumulative distribution function evaluation, tractable under the circular approximation. Assuming independence across agents at timestep $k$, consistent with~\eqref{eq:facdis}, the joint collision probability is
\begin{equation}
\label{eq:joint_collision}
\hat{P}_{\mathrm{joint}}\!\left(\mathbf{Y}_{k \mid t}\right)
=
1 -
\prod_{a \in \mathcal{A}}
\left(
1 - \hat{P}\!\left(\mathbf{y}^a_{k \mid t}\right)
\right).
\end{equation}
To evaluate~\eqref{eq:joint_collision} efficiently across all MPPI samples, we use a parallelized Monte Carlo approximation~\cite{25-dra}. The resulting risk cost is
\begin{equation}
\label{eq:collision_cost}
\mathcal{C}_{\mathrm{risk}}
=
\omega_{\mathrm{soft}} \,
\hat{P}_{\mathrm{joint}}\!\left(\mathbf{Y}_{k \mid t}\right)
+
\omega_{\mathrm{hard}}\,
\mathbf{1}_{\delta}\!\left(
\hat{P}_{\mathrm{joint}}\!\left(\mathbf{Y}_{k \mid t}\right)
\right),
\end{equation}
where $\omega_{\mathrm{soft}}$ is a continuous penalty biasing the planner toward lower-risk trajectories, and $\omega_{\mathrm{hard}}$ is a hard penalty triggered when the collision probability exceeds a threshold $\delta$. The stage cost for rollout $i$ at timestep $k$ decomposes into three components:
\begin{equation}
\label{eq:stage_cost}
S_{k \mid t}^{(i)}
=
\mathcal{C}_{\mathrm{tracking}}
+ \mathcal{C}_{\mathrm{speed}}
+ \mathcal{C}_{\mathrm{risk}},
\end{equation}
where $\mathcal{C}_{\mathrm{tracking}}$ and $\mathcal{C}_{\mathrm{speed}}$ depend only on the ego trajectory, and $\mathcal{C}_{\mathrm{risk}} = \mathcal{C}_{\mathrm{risk}}(\mathbf{Y}_{k\mid t})$ is the stochastic risk cost from~\eqref{eq:collision_cost}. Substituting~\eqref{eq:stage_cost} into~\eqref{eq:approx_inner_exp} gives the total trajectory cost

\begin{equation}
\label{eq:total_cost}
S^{(i)}
=
\phi\!\left(\mathbf{x}^{(i)}_{T\mid t}\right)
+
\sum_{k=0}^{T-1}\sum_{b=1}^{B}
w^{(i,b)}_{k\mid t}\,
S_{k \mid t}^{(i,b)},
\end{equation}
where $S_{k\mid t}^{(i,b)} = \mathcal{C}_{\mathrm{tracking}} +
\mathcal{C}_{\mathrm{speed}} + \mathcal{C}_{\mathrm{risk}}
(\mathbf{Y}^{(i,b)}_{k\mid t})$ denotes the stage cost evaluated against
hypothesis $b$, and the deterministic components are shared across
hypotheses.The planning objective~\eqref{eq:active_objective} is then
approximated by the Monte Carlo estimator of the outer expectation over the $M$ rollouts:
\begin{equation}
J_t
\approx
\frac{1}{M}\sum_{i=1}^{M} S^{(i)}.
\end{equation}
 The MPPI update~\eqref{eq:mppi_update} then maps these costs into importance weights and an updated control sequence.
\begin{rmk}
    \textit{ This solution 
implicitly incentivizes the planner to reduce interaction uncertainty.
The stage-wise MoG surrogate in~\eqref{eq:gmm} approximates how predictive
uncertainty evolves over the planning horizon under a given ego trajectory. This directly affects the collision risk cost in~\eqref{eq:collision_cost}. A diffuse surrogate with competing modes spreads probability mass over a larger region of the agent state space. Since the surrogate is ego-conditioned, ego behaviors that disambiguate agents' intentions, such as assertively
occupying a lane to elicit a clear yield response, tend to
concentrate mixture weights on fewer, more consistent hypotheses, reducing
the estimated risk. 
}
\end{rmk}

\begin{rmk}
    \textit{To reduce the cost of conditioning the predictor on all MPPI samples, we cluster the sampled control sequences via k-means and condition the model on one representative trajectory per cluster, sharing the resulting rollouts within each cluster. This assumes predicted agent behavior varies slowly across similar ego trajectories.}
\end{rmk}

\begin{algorithm}[t]
\caption{Active Interaction- and Risk-Aware MPPI}
\label{alg:ia_mppi}
\begin{algorithmic}[1]

\Initialize{
Nominal control sequence $\mathbf{U}_{t} \gets \mathbf{0}$ \\
MPPI sample count $M$, behavior rollouts $B$
}

\While{task not completed}

\State Observe current ego state $\mathbf{x}_{t}^{\mathrm{meas}}$ and context $\mathcal{H}_t$

\For{$i = 1 \dots M$} \Comment{MPPI rollouts (parallel)}

    \State $\mathbf{U}^{(i)}_{t} \gets \mathbf{U}_{t} + \boldsymbol{\epsilon}^{(i)}, \boldsymbol{\epsilon}^{(i)} \sim \mathcal{N}(\mathbf{0}, \boldsymbol{\Sigma}_{\epsilon})$

    \State  $\mathbf{X}^{(i)}_{0:T \mid t} \gets $ \textit{getRolloutTrajectories}$(\mathbf{U}^{(i)}_{t})$  \hfill \eqref{model}

    \For{$b = 1 \dots B$} \Comment{Behavior rollouts (parallel)}


        \For{$k = 1 \dots T$}

            \State $\begin{aligned}&\mathbf{y}^{(i,b,a)}_{k \mid t},  \pi^{(i,b,a)}_{k \mid t}  \gets  \textit{GARPMSampling}(\\ & \mathbf{Y}^{(i,b)}_{k-1 \mid t}, \mathbf{X}^{(i)}_{k-1 \mid t}, \mathcal{H}_t ) \ \forall  a \in A  \end{aligned} $   
              \eqref{eq:facdis}

        \EndFor
    \EndFor

    \For{$k = 0 \dots T-1$}

        \State $\tilde{p}\!\left(\mathbf{y}^{(a)}_{k\mid t} \mid \cdot \right) \gets$ \textit{getMoG}$(\mathbf{y}^{(i,b,a)}_{k \mid t}, \pi^{(i,b,a)}_{k \mid t} )$ \eqref{eq:gmm}
        \State $\hat{P}_{\mathrm{joint}}\!\left(\mathbf{Y}_{k \mid t}\right) \gets$ \textit{getJointCP}$(\tilde{p}(\mathbf{y}^{(a)}_{k\mid t} \mid \cdot) )$ \hfill \eqref{eq:joint_collision}

        \State $S^{(i)}_{k \mid t} \gets \textit{getStageCost} (\hat{P}_{\mathrm{joint}}, \mathbf{X}^{(i)})$ \hfill 
        \eqref{eq:stage_cost}

    \EndFor

    \State $S^{(i)} \gets $ \textit{getTotalCost}$(S^{(i)}_{k \mid t} )$ \hfill  \eqref{eq:total_cost}

\EndFor

\State $\omega^{(i)} \gets$ \textit{importanceSampling}$(S^{(i)})$
\hfill  \eqref{eq:mppi_update}

\State Update control sequence
$\mathbf{U}_{t}^{*} \gets \sum_{i=1}^{M} \omega^{(i)} \mathbf{U}^{(i)}_{t}
$

\State Apply control $\mathbf{u}^{*}_{0 \mid t}$, Time-shift $\mathbf{U}_{t} \gets \textsc{Shift}(\mathbf{U}_{t}^{*})$

\EndWhile
\end{algorithmic}

\end{algorithm}

\begin{figure*}[!tp]
\centering
\begin{subfigure}{0.30\textwidth}
    \centering
    \includegraphics[width=\linewidth]{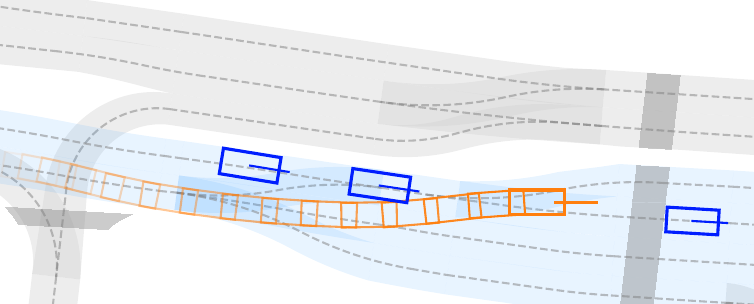}
    \caption{Ours}
\end{subfigure}\hfill
\begin{subfigure}{0.30\textwidth}
    \centering
    \includegraphics[width=\linewidth]{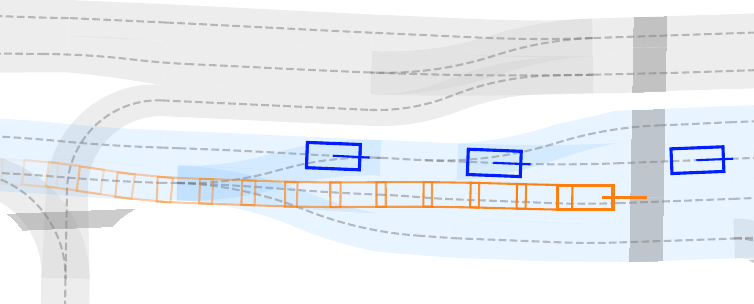}
    \caption{Baseline 1}
\end{subfigure}\hfill
\begin{subfigure}{0.30\textwidth}
    \centering
    \includegraphics[width=\linewidth]{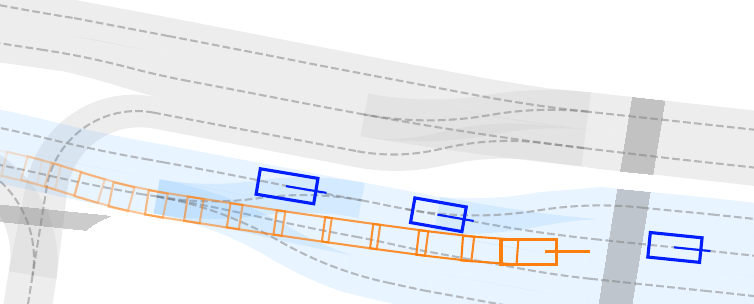}
    \caption{Baseline 2}
\end{subfigure}
\caption{Qualitative comparison of lane merging behavior in the nuPlan simulation environment. The ego-vehicle (orange) interacts with surrounding vehicles (blue) under different approaches. Our method (a) successfully completes the merge by actively probing and inferring the cooperativeness of neighboring vehicles, while the baselines (b–e) remain conservative or fail to identify a feasible merging opportunity.}
\label{fig:scenario1}
\end{figure*}

\begin{figure}[h]
    \centering

    \begin{subfigure}{0.48\columnwidth}
        \centering
        \includegraphics[width=0.9\linewidth]{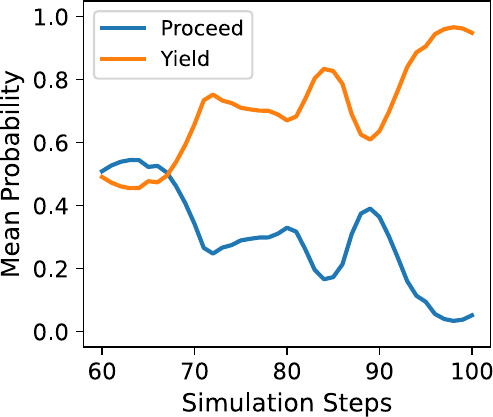}
        \caption{Ours (cooperative case)}
        \label{fig:probing_probs_merge_coop}
    \end{subfigure}
    \hfill
    \begin{subfigure}{0.48\columnwidth}
        \centering
        \includegraphics[width=0.9\linewidth]{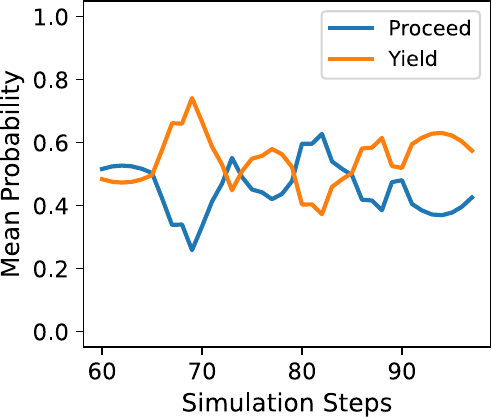}
        \caption{Baseline 1 (cooperative case)}
        \label{fig:probing_probs_merge_coop_no_ego_conditioning}
    \end{subfigure}

    \vspace{0.3cm}

    \begin{subfigure}{0.48\columnwidth}
        \centering
        \includegraphics[width=0.9\linewidth]{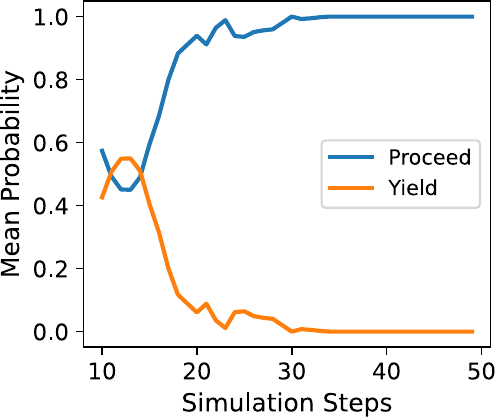}
        \caption{Ours (non-cooperative case)}
        \label{fig:probing_probs_merge_non-coop}
    \end{subfigure}
    \hfill
    \begin{subfigure}{0.48\columnwidth}
        \centering
        \includegraphics[width=0.9\linewidth]{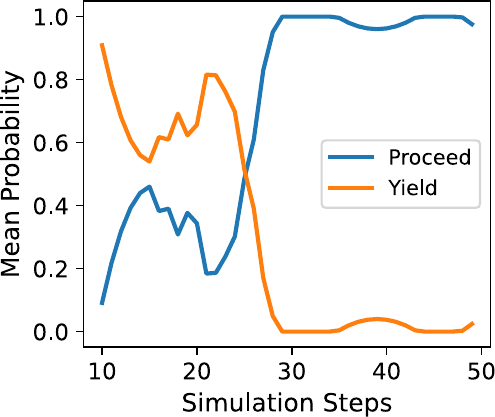}
        \caption{Baseline 1 (non-cooperative case)}
        \label{fig:probing_probs_merge_non_coop_no_ego_conditioning}
    \end{subfigure}

    \caption{Comparison of predicted intent probabilities for cooperative (a,b) and non-cooperative (c,d) cases: An GARPM with and without ego-conditioning run in parallel; only the ego-conditioned feeds our framework. Plots show the ego-conditioned model resolves intent ambiguity of the interacting agent significantly faster. }
    \label{fig:merge_prob}
\end{figure}

\begin{figure}[h]
    \centering

    \begin{subfigure}{0.7\columnwidth}
        \centering
        \includegraphics[width=0.9\linewidth]{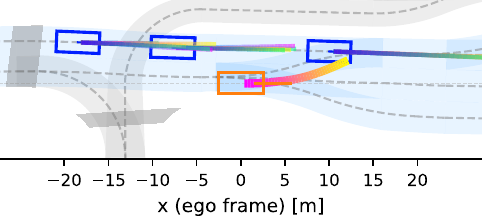}
        \caption{Probing sample predictions}
        \label{fig:sub1}
    \end{subfigure}

    \vspace{2mm}

    \begin{subfigure}{0.7\columnwidth}
        \centering
        \includegraphics[width=0.9\linewidth]{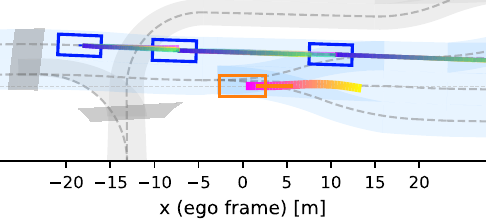}
        \caption{Non-probing sample predictions}
        \label{fig:sub2}
    \end{subfigure}

    \caption{Comparison of ego-conditioned predictions for probing and non-probing MPPI rollouts in an adjacent-lane merge. The probing rollout adopts an assertive trajectory that elicits a yielding response, reducing uncertainty and increasing the inferred probability of cooperative behavior. In contrast, the non-probing rollout remains passive, weakly influencing the interaction and yielding more ambiguous intent predictions.}
    \label{fig:probing_non_probing_predictions}
\end{figure}

\begin{figure*}[!htb]
\centering

\begin{subfigure}{0.30\textwidth}
    \centering
    \includegraphics[width=\linewidth, trim=0 65 0 80, clip]{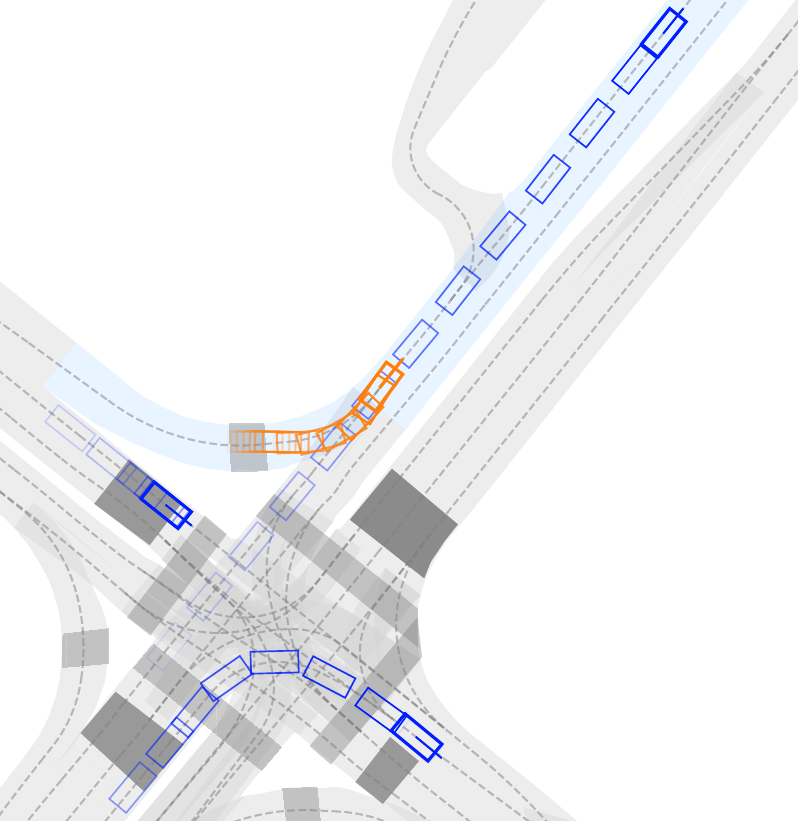}
    \caption{Ours}
\end{subfigure}\hfill
\begin{subfigure}{0.30\textwidth}
    \centering
   \includegraphics[width=\linewidth, trim=0 65 0 80, clip]{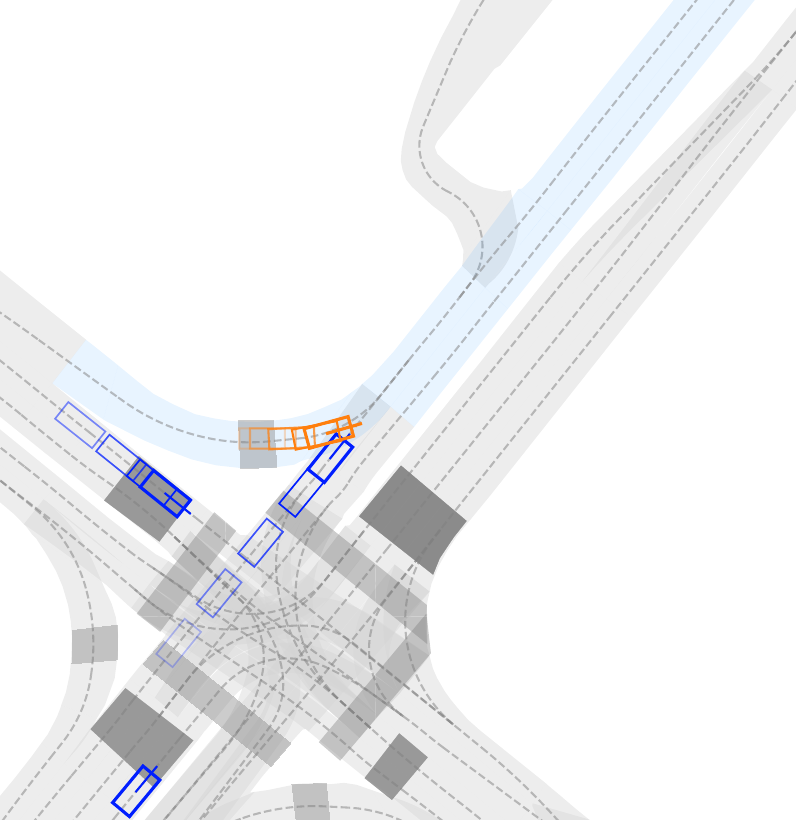}
    \caption{Baseline 3}
\end{subfigure}\hfill
\begin{subfigure}{0.30\textwidth}
    \centering
    \includegraphics[width=\linewidth, trim=0 65 0 80, clip]{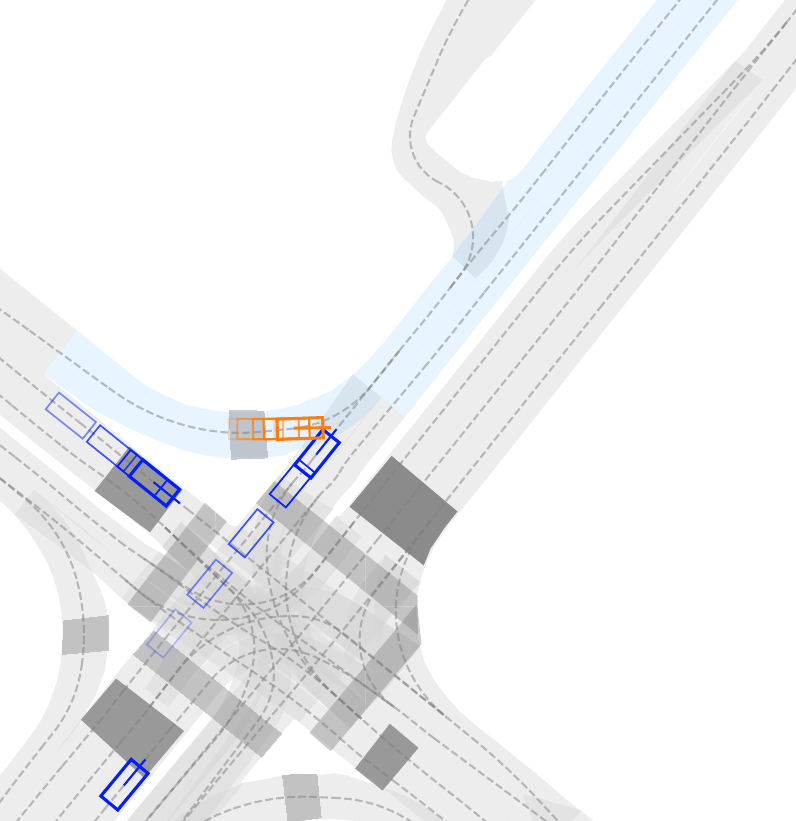}
    \caption{Baseline 4}
\end{subfigure}

\caption{Qualitative comparison of on-ramp merge behavior in the nuPlan simulation environment for a \emph{non-cooperative} approaching vehicle. Our method successfully completes the merge by actively probing and inferring the cooperativeness of neighboring vehicles, while the baselines 3 and 4 are over-confident that the approaching vehicle will yield for it, resulting in a collision.}
\label{fig:scenario2}
\end{figure*}



\section{Results}
We evaluate the proposed approach in interactive urban driving scenarios using the nuPlan simulator \cite{nuplan}. We selected three scenario types in which interaction plays a critical role. For each scenario, we conduct 20 runs with agent cooperativeness levels varied across runs using the MR-IDM model \cite{mr-idm}, spanning cooperative to non-cooperative behaviors.
Four baselines are introduced. Rather than reimplementing the
original methods, all baselines are built upon the same
backbone using SMART as the prediction model and MPPI as
the planner, ensuring fair comparison by isolating the effect
of each design decision and avoiding confounding factors
from architectural differences. Baselines 1–3 correspond to
paradigms established in prior work, while Baseline 4 has
no counterpart in the literature and constitutes a pure
ablation.\\
\textbf{Baseline 1: Multi-modal Predict-Then-Plan} \cite{25-dra}. Agent trajectories are predicted independently of the ego plan and subsequently used for planning.\\
\textbf{Baseline 2: Receding-Horizon Ego-Conditioning} \cite{Chen2023PPADII}. Predictions are conditioned only on the previously planned optimal ego trajectory, without considering alternative future ego behaviors during planning.\\
\textbf{Baseline 3: Unimodal Ego-Conditioned Planning} \cite{dtpp, iann-mppi}. Predictions are conditioned on ego samples but restricted to a single maximum-likelihood trajectory per agent, removing multi-modal uncertainty reasoning.\\
\textbf{Baseline 4: Passive Ego-Conditioned MPPI}. Predictions are conditioned on ego rollouts, but probabilities remain fixed over the horizon, preventing active uncertainty reduction.\\
\vspace{-\baselineskip}
\subsection{Scenario 1: Merging in Adjacent Lane}

In this scenario, the ego vehicle merges into an adjacent lane occupied by vehicles. Since the intent of surrounding agents is not directly observable, the ego vehicle must infer whether nearby vehicles will yield or maintain speed before committing to the merge. 
Qualitative results in Fig.~\ref{fig:scenario1} highlight the difference between our approach and the baselines. The proposed approach successfully completes the merge by actively eliciting responses from neighboring vehicles, while the baselines either remain overly conservative or fail to identify a feasible merge opportunity. Table~\ref{tab:merge_results} shows that the proposed method achieves the highest merge success rate while reducing merge time compared to all baselines. The predict-then-plan baseline remains overly conservative because it ignores ego influence on surrounding agents. Ego-conditioned baselines improve performance but remain limited by either the absence of multi-modality or the inability to exploit uncertainty evolution during planning. Fig.~\ref{fig:merge_prob} illustrates the evolution of the predicted \emph{yield} and \emph{proceed} probabilities for both cooperative and non-cooperative interactions, comparing our ego-conditioned predictions against unconditioned baseline approaches, e.g., Baseline 1. In cooperative scenarios, probing trajectories progressively shift probability mass toward the \emph{yield} mode, enabling the planner to confidently commit to the merge. In contrast, for non-cooperative scenarios, the \emph{proceed} mode rapidly becomes dominant. This dominant proceed mode increases the estimated joint collision probability $\hat{P}_{\text{joint}}(\mathbf{Y}_{k|t})$ defined in \eqref{eq:joint_collision}. This heavily penalizes the trajectory through the risk cost $\mathcal{C}_{\text{risk}}$ in the planning objective \eqref{eq:collision_cost}, naturally steering the MPPI planner to select safer, evasive maneuvers.
The figure also highlights the importance of ego-conditioning: without conditioning on the ego vehicle’s candidate future actions  as formulated in \eqref{eq:ego_cond_pred}, the predicted mode probabilities remain ambiguous throughout the interaction. This limits active uncertainty reduction and results in overly conservative behavior. Fig.~\ref{fig:probing_non_probing_predictions} compares probing and non-probing MPPI rollouts, of our method showing that assertive probing trajectories generate more informative responses and significantly reduce uncertainty regarding surrounding agents’ intent.


\begin{table}[t]
\caption{Adjacent-lane merging results over 20 runs (mean (std)). Time refers to the time the ego-vehicle takes to complete the merge, MR is the merge rate, and Dist. is the longitudinal distance it keeps to the target vehicle after the merge.}
\centering
\resizebox{\columnwidth}{!}{
\begin{tabular}{lcccc}
\toprule
\textbf{Approach} & \textbf{Vel. [m/s]} & \textbf{Time [s]} & \textbf{MR [\%]} & \textbf{Dist. [m]} \\
\midrule
Ours        & 3.39 (0.08) & \textbf{11.80 (0.98)} & \textbf{75.00\%} & 5.85 (2.45) \\
Baseline 1  & \textbf{3.58 (0.11)} & 13.03 (1.34) & 37.50\%  & \textbf{9.22 (2.19)} \\
Baseline 2  & 3.43 (0.09) & 12.53 (0.83) & 56.25\% & 6.42 (2.93) \\
Baseline 3  & 3.38 (0.08) & 12.46 (0.91) & 62.50\% & 6.31 (2.48)\\
Baseline 4  & 3.41 (0.06) & 12.33 (0.73) & 56.25\% & 6.45 (2.84) \\
\bottomrule
\end{tabular}
}
\label{tab:merge_results}
\end{table}

\vspace{-1.5pt}
\subsection{Scenario 2: On-Ramp Merge}

In this scenario, the ego vehicle merges from an on-ramp into a lane occupied by an approaching vehicle, resulting in an interaction in which the ego must infer whether the oncoming vehicle will yield or maintain its speed. 
Fig.~\ref{fig:scenario2} shows a non-cooperative case where the approaching vehicle does not yield. The proposed method correctly infers this behavior and delays merging until the lane is safe, while ego-conditioned baselines overestimate cooperativeness and result in collisions. Table~\ref{tab:ramp_results} shows that the proposed method achieves 0\% collisions while maintaining competitive merge efficiency. 
Despite sharing this exact safety metric, ego-conditioned Baselines 2, 3, and 4 frequently collide (15-25\%) because overly optimistic predictions cause them to dangerously underestimate $\hat{P}_\text{joint}$. Conversely, Baseline 1 consistently overestimates risk, leading to conservative behavior that still fails to reliably anticipate interaction outcomes. 


\begin{table}[t]
\caption{On-ramp merging results over 20 runs (mean (std)). CR refers to the collision rate.}
\centering
\resizebox{\columnwidth}{!}{
\begin{tabular}{lcccc}
\toprule
\textbf{Approach} & \textbf{Vel. [m/s]} & \textbf{Time [s]} & \textbf{CR [\%]} & \textbf{Dist. [m]} \\
\midrule
Ours        & 2.45 (0.59) & 8.28 (1.77) & \textbf{0\%}  & 6.98 (6.22) \\
Baseline 1  & 1.49 (0.52) & 11.5 (1.40) & 10\% & \textbf{14.8 (9.21)} \\
Baseline 2  & \textbf{2.51 (0.59)} & 7.84 (1.87) & 25\% & 7.14 (5.03) \\
Baseline 3  & 2.40 (0.27) & \textbf{7.49 (1.64)} & 20\% & 7.73 (7.84) \\
Baseline 4  & 2.16 (0.24)          & 9.13 (1.46)          & 15 \%   & 7.84 (5.68)          \\

\bottomrule
\end{tabular}
}
\label{tab:ramp_results}
\end{table}

\vspace{-2.1pt}
\subsection{Scenario 3: Unprotected Left Turn}
\begin{figure*}[t]
\centering

\begin{subfigure}{0.30\textwidth}
    \centering
    \includegraphics[width=\linewidth, trim=0 125 0 55, clip]{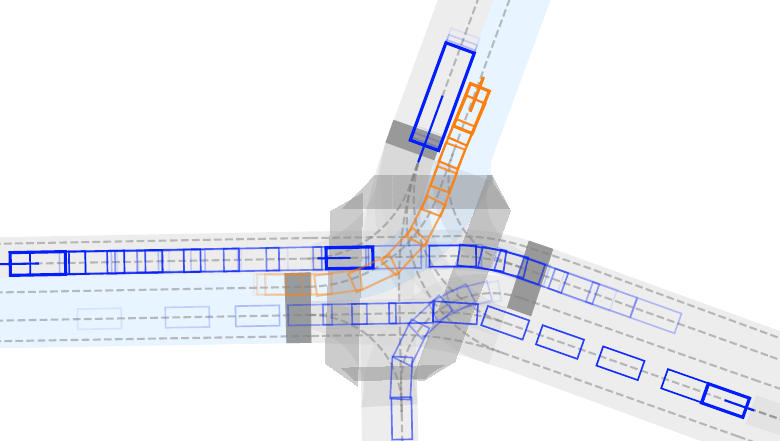}
    \caption{Ours}
\end{subfigure}\hfill
\begin{subfigure}{0.30\textwidth}
    \centering
    \includegraphics[width=\linewidth, trim=0 125 0 55, clip]{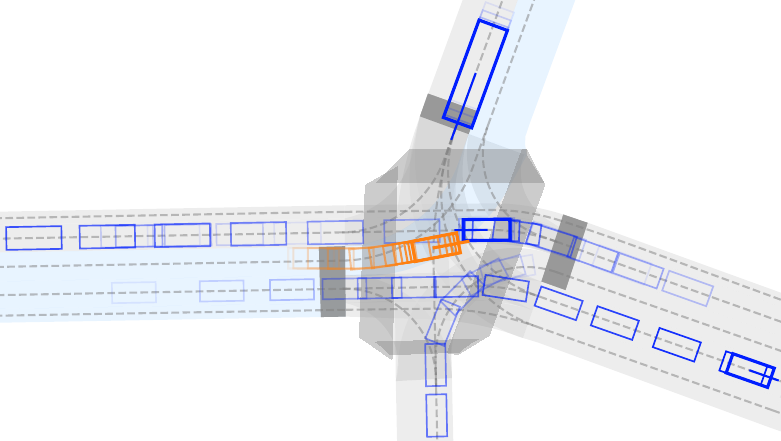}
    \caption{Baseline 2}
\end{subfigure}\hfill
\begin{subfigure}{0.30\textwidth}
    \centering
    \includegraphics[width=\linewidth, trim=0 125 0 55, clip]{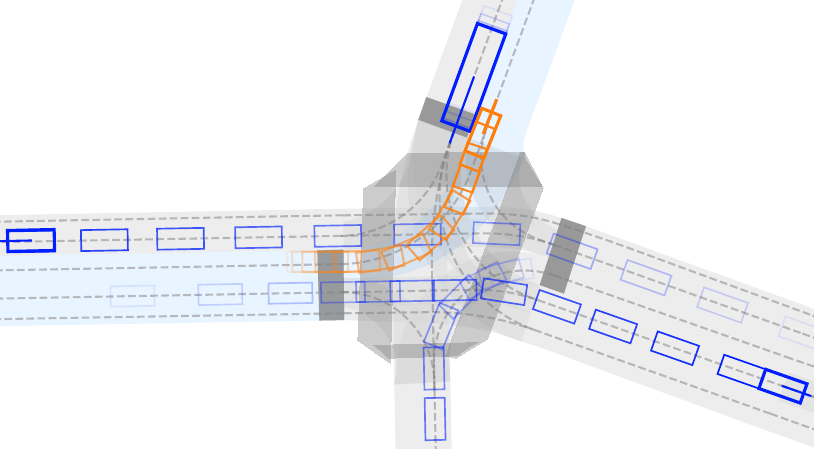}
    \caption{Baseline 3}
\end{subfigure}

\caption{Qualitative comparison of the unprotected left-turn scenario in the nuPlan environment. The proposed approach successfully infers the cooperativeness of the oncoming vehicle through probing interactions and safely executes the left turn. In contrast, Baseline~2 fails to resolve the interaction uncertainty, resulting in a deadlock situation, while Baseline~3 adopts an overly conservative strategy and waits for the approaching vehicle to pass before turning.}
\label{fig:scenario3}
\end{figure*}
In this scenario, the ego vehicle performs an unprotected left turn across dense oncoming traffic. Oncoming vehicles may either maintain speed or yield, requiring the ego vehicle to infer their cooperativeness before committing to the maneuver. 
Qualitative results in Fig.~\ref{fig:scenario3} show that the proposed method successfully infers cooperative behavior and executes the turn efficiently. Baseline 2 fails to resolve interaction uncertainty and results in deadlock situations, while Baseline 3 remains overly conservative and waits unnecessarily. Table~\ref{tab:turn_results} confirms that the proposed approach achieves the best balance between safety and efficiency, obtaining the lowest collision and deadlock rates among all methods.


\begin{table}[t]
\caption{Unprotected left turn results over 20 runs (mean (std)). DL stands for deadlocks}
\centering
\resizebox{\columnwidth}{!}{
\begin{tabular}{lcccc}
\toprule
\textbf{Approach} & \textbf{Vel. [m/s]} & \textbf{Trav. Dist. [m]} & \textbf{CR [\%]} & \textbf{DL [\%]} \\
\midrule
Ours        & \textbf{2.62 (0.36)} & \textbf{39.2 (2.96)} & \textbf{0\%}  & \textbf{5\%} \\
Baseline 1  & 2.60 (0.26) & 38.5 (3.86) & 5\% & 15\% \\
Baseline 2  & 2.34 (0.22) & 34.6 (3.34) & 5\% & 10\% \\
Baseline 3  & 2.51 (0.52) & 35.8 (1.40) & 10\% & 20\% \\
Baseline 4  & 2.32 (0.11) & 34.4 (1.46) & 10\% & 30\% \\
\bottomrule
\end{tabular}
}
\label{tab:turn_results}
\end{table}

\section{Conclusion}
\label{sec:conclusion}
We presented an active interaction-aware motion planning framework leveraging ego-conditioned generative modeling within an MPPI formulation. By conditioning multi-agent trajectory predictions on sampled ego behaviors, the planner explicitly reasons about how its actions influence surrounding agents.  To 
this end, we introduced a tractable nested sampling scheme that approximates the 
resulting double expectation, enabling risk-aware collision probability estimates 
under multi-modal predictive distributions. The resulting formulation allows the 
planner to balance task performance against interaction uncertainty reduction without relying 
on simplified interaction models, explicit belief-space representations, or 
dedicated exploration terms. Simulation results demonstrate that 
our method achieves safer, more efficient, and more decisive behavior compared 
to conventional predict-then-plan and passive interaction-aware baselines.


\bibliographystyle{IEEEtran}
\bibliography{IEEEexample}

\end{document}